\documentclass[conference]{IEEEtran}
\IEEEoverridecommandlockouts
\usepackage{cite}
\usepackage{amsmath,amssymb,amsfonts}
\usepackage{graphicx}
\usepackage{textcomp}
\usepackage[utf8]{inputenc}
\usepackage{algorithm}
\usepackage{algpseudocode}
\usepackage{xcolor}
\usepackage{ulem}
\usepackage{xcolor}
\def\BibTeX{{\rm B\kern-.05em{\sc i\kern-.025em b}\kern-.08em
    T\kern-.1667em\lower.7ex\hbox{E}\kern-.125emX}}
\begin{document}

\title{SPIN: Simulated Poisoning and Inversion Network for Federated Learning-Based 6G Vehicular Networks
}

\author{\IEEEauthorblockN{Sunder Ali Khowaja (Senior Member, IEEE)}
\IEEEauthorblockA{\textit{Department of Telecommunication Engineering} \\
\textit{University of Sindh, Jamshoro, Pakistan}\\
sandar.ali@usindh.edu.pk}
\and
\IEEEauthorblockN{Parus Khuwaja}
\IEEEauthorblockA{\textit{University of Sindh, Jamshoro, Pakistan} \\
parus.khuwaja@usindh.edu.pk}
\and
\IEEEauthorblockN{Kapal Dev (Senior Member, IEEE)}
\IEEEauthorblockA{\textit{Department of Computer Science} \\
\textit{Munster Technological University, Cork, Ireland}\\
kapal.dev@mtu.ie}
\and
\IEEEauthorblockN{Angelos Antonopoulos (Senior Member, IEEE)}
\IEEEauthorblockA{\textit{R\&I Department} \\
\textit{Nearby Computing, Barcelona, Spain}\\
aantonopoulos@nearbycomputing.com}
}

\maketitle

\begin{abstract}
The applications concerning vehicular networks benefit from the vision of beyond 5G and 6G technologies such as ultra-dense network topologies, low latency, and high data rates. Vehicular networks have always faced data privacy preservation concerns, which lead to the advent of distributed learning techniques such as federated learning. Although federated learning has solved data privacy preservation issues to some extent, the technique is quite vulnerable to model inversion and model poisoning attacks. We assume that the design of defense mechanism and attacks are two sides of the same coin. Designing a method to reduce vulnerability requires the attack to be effective and challenging with real-world implications. In this work, we propose simulated poisoning and inversion network (SPIN) that leverages the optimization approach for reconstructing data from a differential model trained by a vehicular node and intercepted when transmitted to roadside unit (RSU). We then train a generative adversarial network (GAN) to improve the generation of data with each passing round and global update from the RSU, accordingly. Evaluation results show the qualitative and quantitative effectiveness of the proposed approach. The attack initiated by SPIN can reduce up to 22\% accuracy on publicly available datasets while just using a single attacker. We assume that revealing the simulation of such attacks would help us find its defense mechanism in an effective manner. 

\end{abstract}


\section{Introduction}
Modern intelligent services demand a communication infrastructure that is both ubiquitous and seamless. Some telecommunication vendors have rolled out the implementation of fifth generation (5G) communication technology that provides a practical estimate of around 20 Gbps of maximum data rate, which is not adequate for end-edge-cloud environments and vehicular network applications that require less than 1ms latency \cite{DBFL, IIFNet}. One of the motivations for moving towards sixth generation (6G) or beyond 5G communication systems is to achieve high reliability and ultra-low latency for realizing future intelligent applications such as autonomous vehicles \cite{6Groad}. Autonomous vehicles and vehicular network applications require the support for high density and scalability in terms of vehicular nodes (end user equipment) along with the governance of sensitive and heterogeneous data in real-time \cite{real-time}. Therefore, it is of utmost importance for applications related to autonomous vehicles to leverage the characteristics of 6G communication networks for realizing intelligent and ubiquitous services.
%Due to the rigorous security requirement, high-data volume, distributed services, and mobile ad-hoc nature, 6G can be considered as a next generation savior for vehicular networks \cite{DBFL}. 
A basic vehicular network comprises of a road-side unit (RSU), a vehicle, and a cloud component that provides the means for communication and provision of intelligent services \cite{DBFL}. For instance, traffic and road side information can be acquired through different signs and boards as well as disseminated to other drivers or vehicles in the city for realizing autonomous driving application. Such service may enhance user experience, smart route planning and navigation, efficient traffic management (intersection control), and safe driving experience (smart cruise control), accordingly. \\
An application scenario of dissemination and sharing lidar or visual data for vehicular networks can be considered. %Such data can be used for sign recognition or object detection to navigate autonomous vehicles in varying traffic conditions. 
Individual vehicles can send the data to a central location for training process and model generation, however, such activity is not only bandwidth inefficient but also makes the individual vehicles vulnerable to several attacks such as leakage of sensitive information, leakage of data, owner information, traffic route, or live access to the vehicle's camera \cite{b5}. In order to avoid such security concerns and improving bandwidth efficiency, federated learning was proposed, especially in vehicular technologies \cite{DBFL}. Individual vehicles can train and generate model with small data quantity, followed by the transmission of the trained model instead of data. This process reduces the bandwidth consumption and security concerns, while compromising on service performance such as recognition accuracy. \\   
Although federated learning provides some security concerning data privacy, the emergence of model inversion attacks that can reproduce the data from model weights (gradient maps), render federated learning methods vulnerable to security threats. Model inversion attacks are studied in a limited way either in general \cite{PrivateAI} or in the context of medical imaging \cite{PGSL}. Such attacks can not only compromise the privacy of users concerning federated learning models but also can affect the overall recognition performance of aggregated model. Existing studies have shown that the defense mechanism for such attacks either involves adding noise, encrypting local parameters, or using late fusion approach \cite{PGSL}. However, designing a defense strategy is as good as knowing what the opponent is capable of. Therefore, designing an efficient simulation method for generating attacks will not only help in improving the defense methods but also will provide an evaluation strategy to test the resiliency of existing methods, accordingly.  \\
To this extent, we propose the concept of simulated poisoning and inversion network (SPIN) attacks for applications concerning vehicular networks. The SPIN framework consists in three basic steps: i) it uses a differential model for generating the data from model weights (gradients); ii) the generated data undergoes adversarial attacks followed by a generative adversarial network (GAN) for generating poisoned data; and iii) a local model is trained and sent to the RSU for aggregation of the global model. We show that after a couple of updates to the global model, the recognition performance drastically decreases, accordingly. To the best of our knowledge, this is the first work to simulate poisoning and model inversion attacks in the context of vehicular networks.\\
The rest of the paper is organized as follows: Section II consolidates some of the works related to model inversion and poisoning attacks. Section III discusses the proposed methodology for SPIN. Section IV presents experimental analysis and results and Section V offers the conclusion of the paper. 
\section{Related Work}
The emergence of advanced technologies concerning communication, deep learning, and computer vision methods have enabled vehicular network applications, including intelligent traffic management, cooperative driving, and autonomous vehicles. Although the management of large-scale data and safety-critical methods have been hot research topics in vehicular networks for quite some time, currently, data and model security are being emerged as point of concerns for the aforementioned applications. This section provides a consolidated review of studies centred towards federated learning and model inversion attacks for vehicular communication networks. \\
Vehicular technology applications face constraints related to data volume, variety, and velocity. Typically, it is estimated that around 30 TB of data is generated by each vehicle each day. These characteristics of vehicular networks are likely to be supported by 5G and beyond networks. Considering the aforementioned facts, the distributed data from the different vehicles need to be utilized in an efficient manner. The study in \cite{Tang2019} provides an in-depth review of comparison between centralized techniques, where the data is sent to a central location (cloud or RSU) for model generation and training, and distributed techniques, which use federated learning that sends local models to the cloud or RSU for model aggregation that enhances the learning accuracy while preserving data privacy. The study suggests that the centralized learning exhibits potential threat of leaking vehicle's information that may or may not contain user personal information as well, high learning latency, and high network bandwidth usage. On the contrary, federated learning improves the data privacy preservation, communication efficiency, and learning accuracy \cite{DBFL}. Since then, federated learning has been used extensively by a number of studies for vehicular network applications. Some of the studies focus on improvement of learning accuracy and communication efficiency, while others highlight the problem of security, specifically model inversion attacks. \\
Du et al. \cite{Du2020} discussed the integration of federated learning with vehicular communication networks while suggesting several challenges and factors that could be considered for improving federated learning-based vehicular networks. Ye et al. \cite{Ye2020} consider the problem of 3D object detection for vehicular network application. Their approach is about aggregating selective models through two-dimension contract theory that eventually improves the interaction between vehicular nodes and RSU. Brik et al. \cite{Brik2020} identify several challenges, open issues, and future directions concerning federated learning and wireless networks of unmanned aerial vehicles. Although some of the challenges resonate with the conventional vehicular networks, some others were specific to unmanned aerial vehicles such as drones and air taxis. He et al. \cite{He2020} focus on the energy-efficiency aspect of the vehicular communication networks in conjunction to federated learning. The study proposes resource allocation algorithms along with importance-aware joint data selection to improve the efficiency as well as the learning speed. \\
Some of the studies (such as \cite{Yu2021}) have considered the security and privacy preservation aspect of federated learning-based vehicular networks by proposing collaborative data leakage protection and content popularity caching methods. However, these studies focus only on the data privacy preservation. A few works have worked on the defense against model inversion attacks with respect to federated learning networks. The idea of model inversion attacks is to reconstruct the sensitive features of the training set from gradients of the model \cite{PrivateAI, PGSL}. Model inversion attacks gained popularity from a study deep leakage from gradients \cite{DLG} that show how the reconstruction can be carried out. To cope with such attacks, some studies consider adding noise to the input data \cite{Li2022}, and some consider performing intentional initialized attacks to the input data \cite{PGSL}. Recently, Khowaja et al. have proposed PGSL \cite{PGSL} and Private AI \cite{PrivateAI} networks that initialize intentional attacks to the input data as a defense against model inversion attacks. Papernot et al. \cite{PATE} have proposed the method private aggeregation of teacher ensembles (PATE) that adds noise to the labels for defense against model inversion attacks. Pan et al. \cite{FLPATE} have extended their work to be applied in the context of federated learning approach for preventing model inversion attacks. Chen et al. \cite{Chen2022} have proposed the use of homomorphic encryption and blockchain technologies to preserve the deep learning model's privacy. A study in \cite{Zhang2019} has used GANs to mimic samples from training data in a federated learning-based network by creating a replica of global model as discriminator. Their idea is to train the GAN network based on the updates of the global model in order to better generate the training samples. The problem with this method is that it assumes the acquisition of global model and the training data to begin with. This study is unique in the sense that it does not need training data and the proposed network will induce the model poisoning attack in a subtle way, i.e. slow poisoning. To the best of our knowledge, none of the works provided a systematic way to simulate the model inversion attacks and gradually corrupt the decision systems through model poisoning in the context of federated learning-based vehicular network applications.  

\begin{figure*}[htbp]
\centerline{\includegraphics[width=\linewidth]{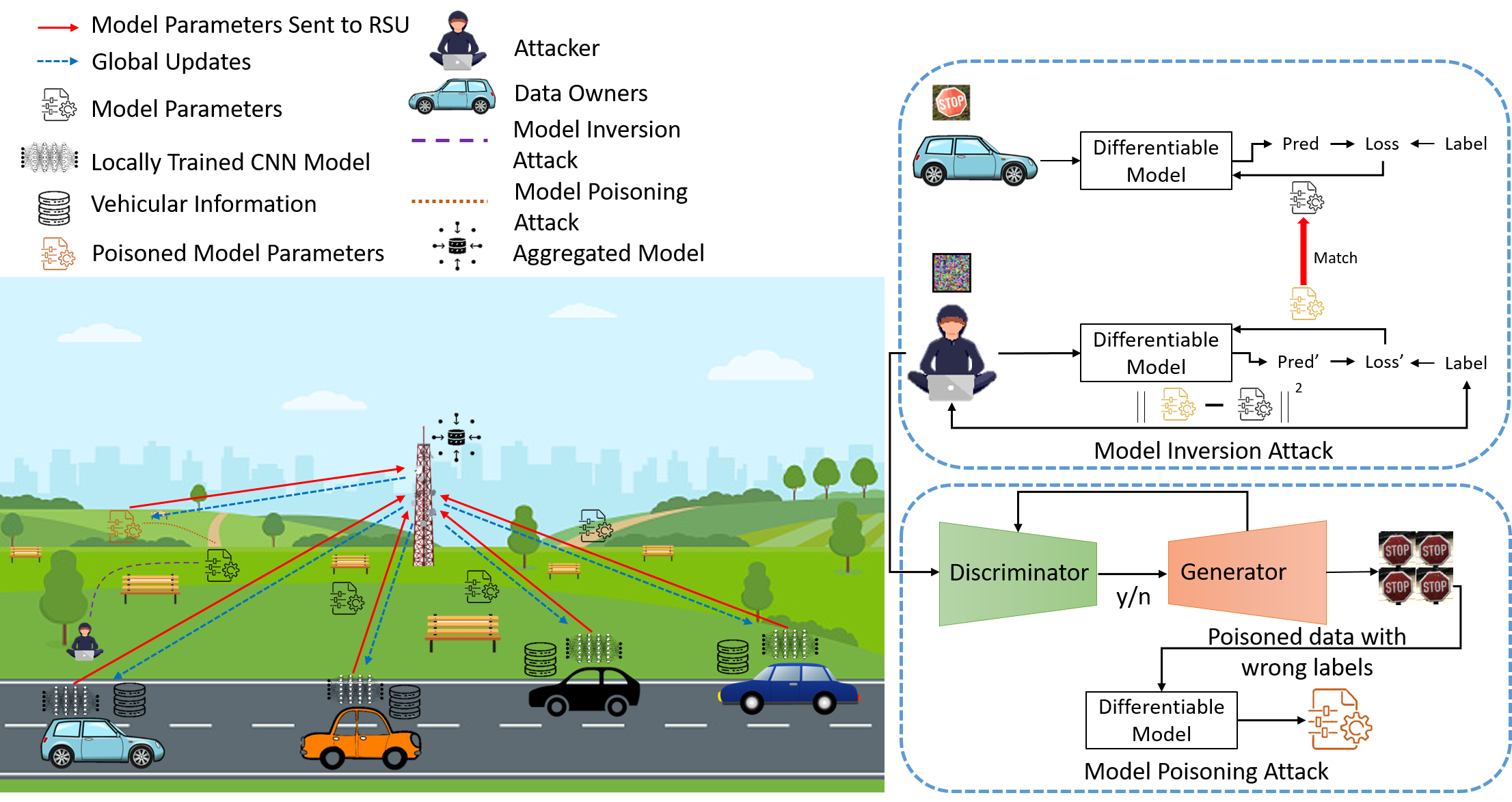}}
\caption{Proposed SPIN architecture for simulating model inversion and poisoning attacks in Federated Learning-based Vehicular networks}
\label{SPIN}
\end{figure*}

\section{Simulated Poisoning and Inversion Network (SPIN)}
The proposed SPIN architecture for simulating attacks is shown in Figure \ref{SPIN}. The scenario depicts that convolutional neural network (CNN) models are trained with vehicular and traffic information data at data owners side. The model parameters are then sent to the RSU for further model aggregation and updated model is sent back to the data owners (vehicles) through the global update. In addition, the scenario also depicts that an attacker can intercept model parameters and initiate model inversion attack to reconstruct the data, accordingly. The estimated reconstructed data is then sent to generative adversarial networks (GANs) to generate similar data with added poisoning to the pixels and labels, accordingly. A model is trained on the poisoned data and is sent to the RSU as one of the model parameters for aggregation. The global update will also be received by the attacker and with every update the discriminator is updated which makes the generator more stronger and resilient, hence, with passing updates, the model poisoning attack will start disrupting the model's performance for all the vehicles that will apparently receive the updates from the RSU. We provide the details for the model inversion attack and model poisoning attack modules in the subsequent subsections, respectively. 

\subsection{Model Inversion Attack}
For applying model inversion attack, we consider the deep leakage from gradient \cite{DLG} method as a reference point. We assume that the training is performed with standard synchronous distributed process. Let us denote the differentiable model from the data owner as $\mathbf{M}(x,W)$, where $x, W$ refer to the training data and parameter weights, respectively. Each vehicular node $p$, at each step $i$ computes the gradients from a sampled minibatch $(x_{i,p}, y_{i,p})$ as shown in equation 1.
\begin{equation}
    \nabla W_{i,p} = \frac{\partial \mathbf{L}(\mathbf{M}(x_{i,p}, W_{i}), y_{i,p})}{\partial W_{i}}
\end{equation}
Let us assume that the $\mathbf{Q}$ servers are being used for averaging the gradients. The formulation for updating the weight parameters can be represented as shown in equation 2.
\begin{equation}
    \nabla W_i = \frac{1}{\mathbf{Q}}\sum_{q}^{\mathbf{Q}} \nabla W_{i,q}; \; \; W_{i+1} = W_{i} - \eta \nabla W_{i}
\end{equation}
This study assumes that the attacker can steal a vehicle $\mathbf{V}$'s training data $(x_{i,\mathbf{V}}, y_{i,\mathbf{V}})$ by intercepting its model parameters $\nabla W_{i,\mathbf{V}}$. It should be noted that the assumption of shared $\mathbf{M}()$ and $W_i$ holds true for all synchronized distributed optimization processes. The first step for recovering data from $\mathbf{V}$'s model gradients is to randomly initialize dummy input data and labels $(x',y')$, accordingly. The said dummy data is fed to the model in order to acquire dummy gradients $\nabla W'$ as represented by formulation in equation 3.
\begin{equation}
    \nabla W' = \frac{\partial \mathbf{L}(\mathbf{M}(x', W), y')}{\partial W}
\end{equation}
As suggested earlier, the studies showed that bringing the dummy gradients close to the intercepted gradients through optimization eventually realizes dummy data close to the real one. Therefore, the optimization function to obtain the reconstructed data by reducing the gap between dummy gradients and original one is shown in equation 4.
\begin{equation}
    \begin{aligned}
    x^{'*}, y^{'*} & = \arg\min_{x',y'} \| \nabla W' - \nabla W \|^2 \\
    & = \arg\min_{x',y'} \| \frac{\partial \mathbf{L}(\mathbf{M}(x', W), y')}{\partial W} - \nabla W \|^2
    \end{aligned}
\end{equation}
There are mainly two assumptions that need to be hold true in order to make the aforementioned optimization function realizable. The first one is that the distance $\| \nabla W' - \nabla W \|^2$ should be differential and the second one is that the optimization function requires 2nd order derivatives to be computed, therefore, $\mathbf{M}$ should be twice differentiable, accordingly. It should be noted that the aforementioned assumptions hold true for the majority of the deep learning models and tasks, especially the ones that employ (convolutional neural networks) \cite{DLG}.

\subsection{Model Poisoning Attack}
The SPIN architecture uses generative adversarial networks (GANs) for generating poisoned data from $(x'*,y'*)$. The GANs employ two neural networks, i.e., generator $\mathbf{G}$ and discriminator $\mathbf{D}$, which are trained in an adversarial fashion. The discriminator network is trained to distinguish between real and fake data while the generator network generates the training data by mimicking discriminative network. A study \cite{Zhang2019} used GANs to mimic samples from training data in a federated learning-based network by creating a replica of global model as discriminator. Thus, the study assumes the acquisition of global model as well as training data to begin with, which this study does not considers. We generate the data by applying model inversion attacks, accordingly. However, once the data is generated, we apply model poisoning by changing the labels. Subsequently, the attacker receives the model updates in the form of global model, which is replicated and used as the discriminator to improve the poisoned data generation process. Therefore, high quality poisoned data can be generated with each passing update. Let us consider a random vector $r$ as an initialization, $\mathbf{M}(x,W)$ as differentiable model that will be initialized as discriminator model $\mathbf{D}$. The adversarial objective function for improving the poisoned data generation process is shown in equation 5.
\begin{equation}
    \begin{aligned}
    \min_{\mathbf{G}}\max_{\mathbf{D}} \rho(\mathbf{D}, \mathbf{G}) & = \mathbb{E}_{r \sim \sigma_{r}(r)}[log(1-\mathbf{D}(\mathbf{G}(r)))] +\\
    & = \mathbb{E}_{x^{'*} \sim \sigma_{data}(x^{'*})}[log \mathbf{D}(x{'*})]
    \end{aligned}
\end{equation}
where $\sigma_{r}(r)$ represents the distribution of random initialization $r$ and $\sigma_{data}(x^{'*})$ refers to the distribution of reconstructed images. The GAN is trained for several iterations until the loss of the objective function achieves Nash equilibrium. The process for the generation of poisoned data is summarized in algorithm 1. 
\begin{algorithm} [ht]
%\DontPrintSemicolon\SetAlgoLined
%\noindent\rule{6.3cm}{0.4pt}

\textbf{Input}: $\mathbf{M}(x,W)$ \\
\textbf{Output}: Trained model using Poisoned Data and labels $\mathbf{M}(x^{\ddagger},W^{\ddagger}), y^{\ddagger}$\\
\emph{Initialize} $\mathbf{G}$ and $\mathbf{D}$\\
%\normalem
\textbf{for}: each epoch $e \in (1, ..., E)$ \textbf{do} \\  
\smallskip \; \;   Initialize $\mathbf{D} \; \leftarrow \; \mathbf{M}(x,W)$ \\
\smallskip \; \;   Run $\mathbf{G}$ for generating $x^{\ddagger}$ \\
\smallskip \; \;   Use $\mathbf{D}$ to Update $\mathbf{G}$ \\
\smallskip \; \;   Assign wrong label to $x^{\ddagger}$ \\
\textbf{end}\\
\textbf{Train} a local model on the poisoned data $\mathbf{M}(x^{\ddagger},W^{\ddagger}), y^{\ddagger}$\\
\textbf{Return} $\mathbf{M}(x^{\ddagger},W^{\ddagger}), y^{\ddagger}$
%\noindent\rule{6.3cm}{0.4pt}
\ULforem
\caption{\label{code:a1} Poisoned Data Generation}
\end{algorithm}

\section{Experimental Evaluation}
\subsection{Datasets}
In this study, we validate our approach on MNIST and GTSRB datasets. Both of the datasets are benchmark in many applications that concern deep learning and both of them are relevant to our study, as one represents numbers while the other represents the traffic signs. MNIST comprises of 70k grayscale images of digits from 0 - 9. The images in MNIST dataset have a fixed size of $28 \times 28$ pixels. MNIST is divided into 60k and 10k training and testing records, respectively. \\
On the other hand, the Gernman Traffic Sign Recognition Benchmark (GTSRB) \cite{GTSRB} comprises of 51,840 color images of 43 traffic signs. The images in GTSRB dataset have varying image sizes, i.e. $15 \times 15$ to $250 \times 250$. In this study, we only consider four traffic signs, i.e. Stop, Do not Enter, 20 Km/h and 120 km/h, respectively. The dataset is splitted into training, validation, and test sets with ratios of 50\%, 25\%, and 25\%, respectively. 

\subsection{Experimental Setup}
The proposed SPIN architecture mainly comprises of a generator and a classifier/discriminator. We use the standard ResNet-56 \cite{ResNet} as the choice of convolutional neural network to train local models, thus, resulting in the classifier/discriminator. The only change we applied is the replacement of activation layer from ReLU to sigmoid as the proposed work assumes that the functions are twice differentiable. Furthermore, we have removed the strides due to the same reason. We have used a deconvolution based network as the generator in this work. The inputs from both datasets were resized to $64 \times 64$ pixels. The images that were lower than this resoultion were upscaled (4x) using RDFDBK model \cite{NTIRE} and then resized to the aforementioned resolution, accordingly. The classifier/discriminator for both datasets are the same, except the last layer due to the varying number of classes. We used the kernel size of $4 \times 4$ with length of 512 for the generator network. We used ReLU as the activation function for the generator network, accordingly.\\
In the proposed work, we set the number of vehicles to 9, along with a single attacker scenario, suggesting that out of 10 participants one would be attacker and 9 would be normal/benign vehicles. For the model inversion attack module, we have used the learning rate of 0.5, number of iterations 10, optimizing iterations 1k, and history size 100 along with L-BFGS method \cite{LBFGS}. For the model poisoning attack part, the benign vehicles will train with learning rate of 0.05 for 4 local epochs while the attacker is trained with the learning rate of 0.01 for 10 local epochs. The learning rate for attacker decays by 20\% after every 3 epochs. All the experiments are carried out using federated learning-based methods for 100 communication rounds. After every iteration, the models from vehicles will be averaged sequentially to construct a global model. We chose PyTorch \cite{Pytorch} framework for our experiments based on federated learning settings. The experiments are carried out on a PC having RTX3060Ti GPU with 32GB of RAM. 

\subsection{Qualitative Effectiveness of Generative Model}
In this subsection, we show the generative effectiveness of the proposed model in a federated learning-based vehicular network system. The number of vehicles in this case is 9, whereas a single attacker scenario is considered. The results for the generated images on MNIST and GTSRB and their comparison with real images are shown in Figure 2. Although we have shown only 4 generated classes from MNIST, it should be noted that this study considers all ten classes for MNIST and only four classes from GTSRB, accordingly. The results show that without accessing the data from start, the proposed method can traverse the data from differentiable model in a satisfactory manner.

\begin{figure}[htbp]
\centerline{\includegraphics[width=0.95\linewidth]{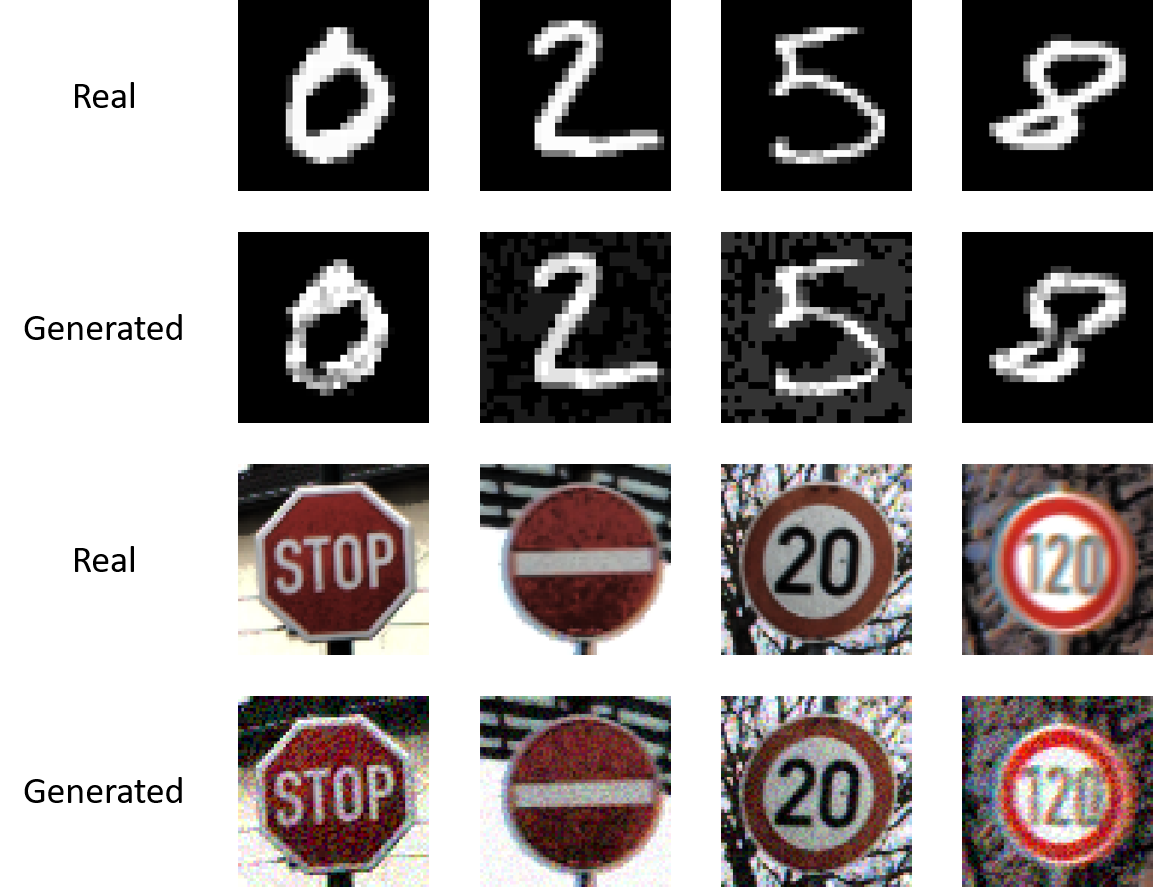}}
\caption{Data Generation by applying model inversion and model poisoning attack on MNIST and GTSRB datasets.}
\label{fig1}
\end{figure}

\begin{figure}[htbp]
\centerline{\includegraphics[width=0.95\linewidth]{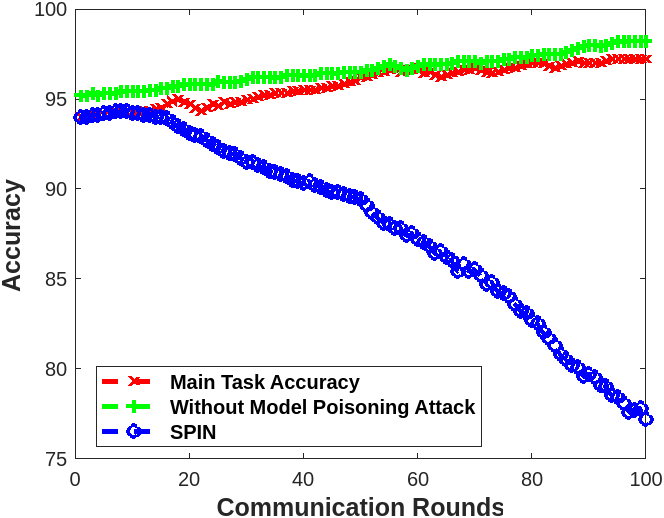}}
\caption{Attack simulation results on MNIST using SPIN}
\label{fig_res_1}
\end{figure}

\begin{figure}[htbp]
\centerline{\includegraphics[width=0.95\linewidth]{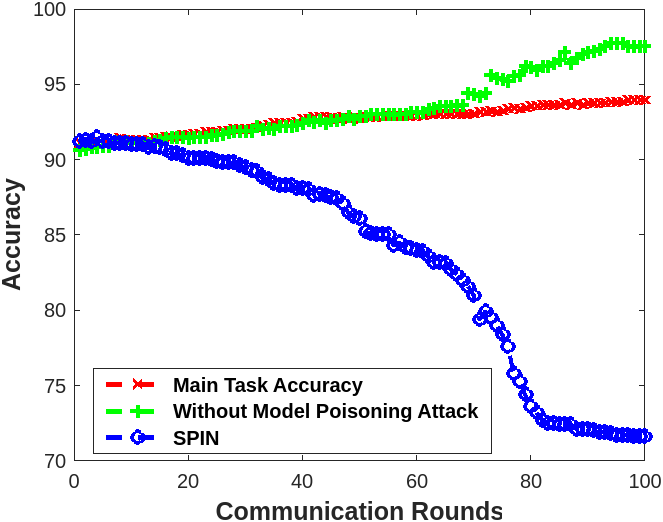}}
\caption{Attack simulation results on GTSRB using SPIN}
\label{fig_res_2}
\end{figure}

\subsection{Experimental Results}
In this study, we define the main task accuracy as the success rate of the recognition task when the model poisoning attack is not initiated. The term without model poisoning attack refers to the addition of model that is trained on the data reconstructed through model inversion attacks. Finally, the SPIN refers to the model inversion and poisoning attack. The experiments are performed on both the MNIST and the GTSRB datasets. The results for both datasets are shown in Figure 3 and 4, respectively. \\
For the main task accuracy, we used the models from 9 benign vehicles that participated in federated learning process. It is shown that the accuracy of around 97\% and 94\% is achieved on both datasets. For the experiment without model poisoning attack, we used 9 benign vehicles and 1 attacker to participate in the federated learning process. In the initial phase of data generation, the accuracy does not improve much for GTSRB dataset, however, as the updates are being sent, the generation process is improved along with the accuracy, which is around 98\% for both the datasets. It should be noted that we did not poison the labels for this experiment to show the effectiveness of the data generation process. Similar trend is noticed for the attack scenario, where 9 benign vehicles and 1 attacker participated in the federated learning process, but, this time, we initiated poisoning attack. It can be noticed that in the beginning, due to not that good generation of samples, the accuracy does not degrade significantly, however with passing of communication rounds and acquiring global updates, the generation process is improved. This leads to significant degradation in accuracy for both the datasets, accordingly. The accuracy decreases up to 19\% and 22\% for MNIST and GTSRB datasets, respectively. We assume that the results for attacks from SPIN architecture are quite effective, considering that only 1 attacker participated in this process. 

\section{Conclusion and Future Work}
This study has proposed a novel simulated poisoning and inversion network (SPIN) for initiating attacks in federated learning-based vehicular applications for beyond 5G and 6G networks. We have shown the effectiveness of the proposed inversion attack module and the generator network to recover the data from differential model. The proposed work has illustrated the efficacy of attacks by mimicking the data from model and then poisoning the labels with respect to federated learning process. It is a kind of a sneak attack that camouflages itself as one of the participant by reconstructing the data from the differential model rather than invading other participant's data for generating poison attack. The assumption in the proposed scenario is quite realistic and is compliant with the emerging threats for current political and security situations. We have also shown that the generation process improves with each passing round and global updates, accordingly. \\
As future work, we plan to conduct more experiments with varying number of attackers and benign vehicles to validate the efficacy of SPIN architecture. Furthermore, we also intend to propose a defense mechanism for such attacks concerning federated learning-based vehicular applications. 

%\section*{Acknowledgment}

%To check

%\section*{References}
% Please number citations consecutively within brackets \cite{b1}. The 
% sentence punctuation follows the bracket \cite{b2}. Refer simply to the reference 
% number, as in \cite{b3}---do not use ``Ref. \cite{b3}'' or ``reference \cite{b3}'' except at 
% the beginning of a sentence: ``Reference \cite{b3} was the first $\ldots$''

% Number footnotes separately in superscripts. Place the actual footnote at 
% the bottom of the column in which it was cited. Do not put footnotes in the 
% abstract or reference list. Use letters for table footnotes.

% Unless there are six authors or more give all authors' names; do not use 
% ``et al.''. Papers that have not been published, even if they have been 
% submitted for publication, should be cited as ``unpublished'' \cite{b4}. Papers 
% that have been accepted for publication should be cited as ``in press'' \cite{b5}. 
% Capitalize only the first word in a paper title, except for proper nouns and 
% element symbols.

% For papers published in translation journals, please give the English 
% citation first, followed by the original foreign-language citation \cite{b6}.

% \vspace{12pt}
% \color{red}
% IEEE conference templates contain guidance text for composing and formatting conference papers. Please ensure that all template text is removed from your conference paper prior to submission to the conference. Failure to remove the template text from your paper may result in your paper not being published.

\end{document}